\documentclass[letterpaper, 10 pt, conference]{ieeeconf}  

\IEEEoverridecommandlockouts                              

\usepackage{graphics} 
\usepackage{epsfig} 
\usepackage{mathptmx} 
\usepackage{times} 
\usepackage{amsmath} 
\usepackage{amssymb}  
\usepackage{hyperref}
\usepackage{multirow}
\usepackage{multicol}
\usepackage{bbding}
\usepackage{colortbl}
\usepackage{booktabs}

\title{\LARGE \bf
REMOTE: Real-time Ego-motion Tracking for Various Endoscopes via Multimodal Visual Feature Learning}

\author{Liangjing Shao$^{1,2}$, Benshuang Chen$^{1,2}$, Shuting Zhao$^{1,2}$ and Xinrong Chen$^{1,2,*}$ 
\thanks{$^{1}$Academy for Engineering \& Technology, Fudan University} %
\thanks{$^{2}$Shanghai Key Laboratory of Medical Image Computing and Computer-Assisted Intervention, Fudan University} %
\thanks{This work was supported by National Natural Science Foundation of China (Grant No.82472116),  Shanghai Natural Science Foundation (Grant No.24ZR1404100) and Key Research and Development Plan of Ningxia Hui Autonomous Region (Grant No. 2023BEG02035).} 
\thanks{* corresponding author: {\tt\small chenxinrong@fudan.edu.cn}} 
}

\begin{document}

\maketitle
\thispagestyle{empty}
\pagestyle{empty}

\begin{abstract}
Real-time ego-motion tracking for endoscope is a significant task for efficient navigation and robotic automation of endoscopy. In this paper, a novel framework is proposed to perform real-time ego-motion tracking for endoscope. Firstly, a multi-modal visual feature learning network is proposed to perform relative pose prediction, in which the motion feature from the optical flow, the scene features and the joint feature from two adjacent observations are all extracted for prediction. Due to more correlation information in the channel dimension of the concatenated image, a novel feature extractor is designed based on an attention mechanism to integrate multi-dimensional information from the concatenation of two continuous frames. To extract more complete feature representation from the fused features, a novel pose decoder is proposed to predict the pose transformation from the concatenated feature map at the end of the framework. At last, the absolute pose of endoscope is calculated based on relative poses. The experiment is conducted on three datasets of various endoscopic scenes and the results demonstrate that the proposed method outperforms state-of-the-art methods. Besides, the inference speed of the proposed method is over 30 frames per second, which meets the real-time requirement. The project page is here: \href{https://remote-bmxs.netlify.app}{remote-bmxs.netlify.app}
\end{abstract}


\section{INTRODUCTION}
Endoscopy has become the primary medical method for the surgery and the examination of the body cavities. Due to the limited visual field and the narrow pathway in the body cavities, the navigation is critical for the safety, accuracy and efficiency of endoscopic treatment, especially automatic robot-assisted endoscopy \cite{nav}.

The common methods for navigation of endoscopy include optical tracking system \cite{opt} or magnetic tracking system \cite{mag}. However, the optical tracking system are easily influenced by the occlusion by human, while the magnetic localization could be interfered by the electromagnetic field from medical instruments. Meanwhile, such tracking systems are expensive and complex to implement. Therefore, the vision-based tracking method has become significant for the navigation of endoscopy.

The vision-based tracking methods can be categorized into two terms, including pose estimation and ego-motion estimation. The target of pose estimation is to estimate the pose of the object from the image of the scene comprising the object, which is impossible for the enclosed narrow body cavities. Differently, the task of ego-motion estimation is to predict the pose of the camera from the observation captured by itself. Therefore, real-time ego-motion tracking is the main method of vision-based tracking for endoscopes. 

The traditional methods for ego-motion estimation are based on feature matching and iterative optimization \cite{tram1}, \cite{tram2}, which is time-consuming and not suitable for real-time tracking. Therefore, learning-based methods are proposed to estimate relative ego-motion between two continuous images and then more efficient ego-motion tracking is performed. The primary pipeline of the relative pose estimation consists of feature extraction and pose decoder \cite{review}. However, currently, most of the existing methods for ego-motion estimation just extract feature representation from two images separately \cite{offset},\cite{borrego} or from the concatenation of two images \cite{endoslam},\cite{liu}. Moreover, the structures of the existing pose decoders are mainly based on a fully-connected layer \cite{yang} or a simple network consisting of four convolution layers \cite{liu}, which are not enough for extracting more complete feature representation from the feature maps. In some researches, recurrent neural networks (RNNs) are applied to extract temporal feature from the sequence of endoscopic images \cite{walch},\cite{rnnm},\cite{borrego},\cite{li}. However, the training of RNNs needs a large quantity of the whole endoscopic videos, which is expensive for establishment of the dataset.

\begin{figure}
    \centering
    \includegraphics[width=8.5cm]{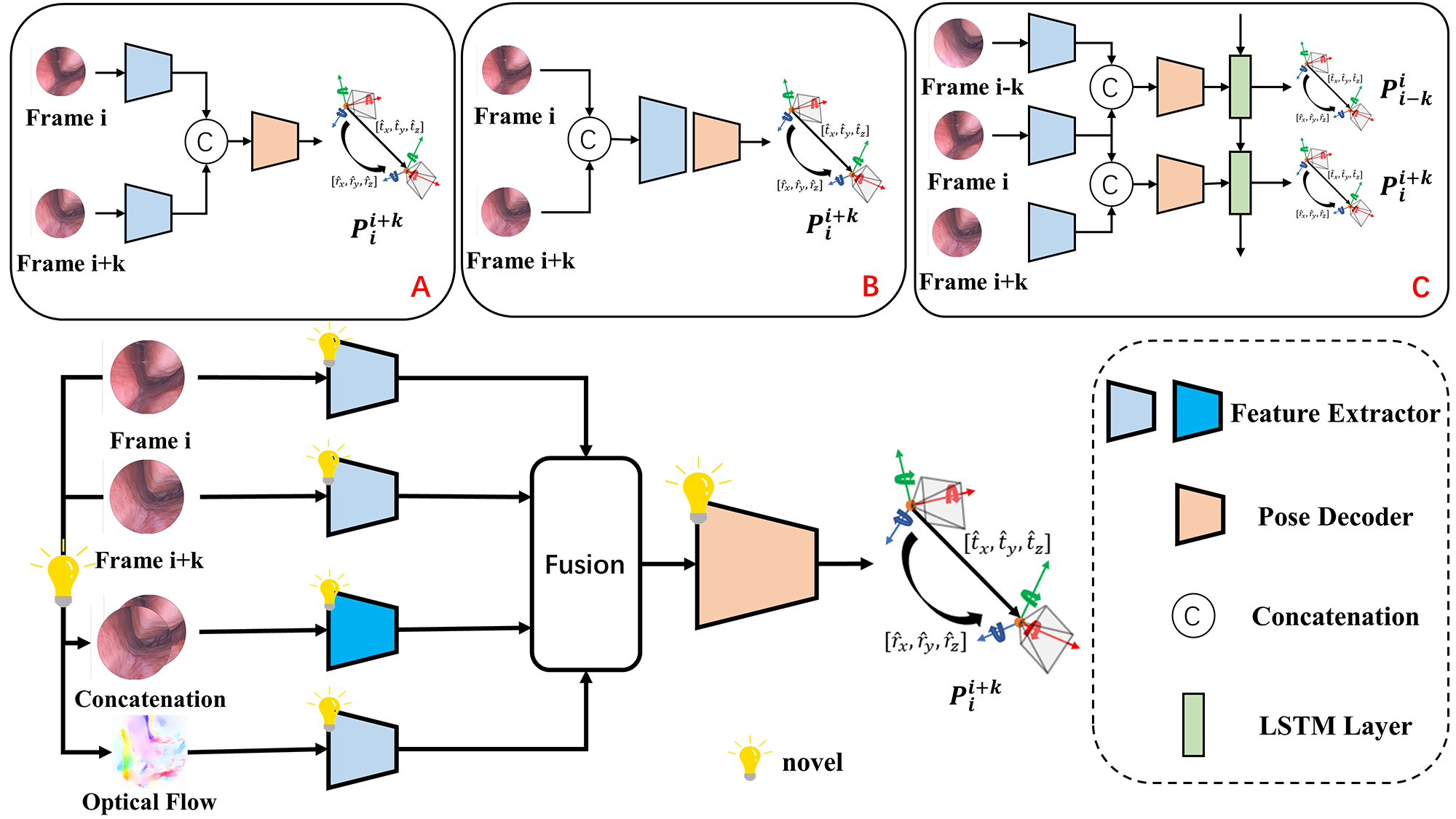}
    \caption{The pipelines of the proposed network and previous methods (categorized into the structures of A, B and C) to estimate relative pose of endoscopes. \textbf{A}: feature of adjacent observations are extracted separately and then concatenated \cite{simcol},\cite{offset}. \textbf{B}: the feature is extracted from the concatenation of adjacent observations \cite{endoslam},\cite{liu},\cite{shao}. \textbf{C}: based on A, LSTM layers are used in the end \cite{borrego},\cite{li}.}
    \label{head}
\end{figure}

To this end, a novel deep learning-based framework is proposed to perform real-time ego-motion tracking for endoscope. In the proposed network, multi-modal visual features are extracted for relative pose estimation, as Fig. \ref{head} shows. The main contributions of our work can be summarized as the following:
\begin{enumerate}
    \item A multi-modal visual feature learning framework is proposed to predict relative pose transformation based on the current observation and the previous frame.
    \item Due to more information in the channel dimension of the concatenated image , a novel feature extractor based on attention mechanism is proposed to extract and integrate the multi-dimensional information from the concatenation of two continuous frames.
    \item To extract more complete feature representation from the feature map, a novel pose decoder is designed to predict the vector of relative pose from the fusion of the features.
    \item The proposed framework is compared with existing state-of-the-art methods on three different datasets. The experimental results demonstrate that the proposed framework performs the most accurate ego-motion tracking. Moreover, the inference speed of the proposed framework meets the real-time requirement.
\end{enumerate}

\section{Related Work}
Most of the existing methods for ego-motion tracking are based on the structure of PoseNet\cite{posenet}, which consists of a feature extractor and a pose decoder. Some works utilize different feature extractors and pose decoders in the similar structure. Based on PoseNet, Nasser et al. \cite{naseer} utilize VGG instead of GoogLeNet and add two fully-connected layers into the pose decoder to extract more complete feature extraction. Due to the excellent performance of ResNet, Wang et al. \cite{wang} and Bui et al. \cite{bui} respectively use ResNet-18 and ResNet-34 as feature extractors in their frameworks. The similar structure is also used in the most of previous methods for ego-motion estimation of endoscope \cite{offset},\cite{simcol},\cite{endoslam},\cite{borrego},\cite{liu}. Differently, multi-modal visual features are extracted in the proposed work. Moreover, a novel feature extractor for multi-dimensional feature integration is proposed and a novel pose decoder to extract more complete feature representation is designed. 

Recently, attention mechanisms are widely applied in the ego-motion estimation of endoscope. For instance, Zhou et al. \cite{zhou} apply the attention mechanism into the pose decoder to extract geometrically robust feature from the feature map. To extract more contextual feature, Shavit et al. \cite{shavit} directly replace CNN networks with a Transformer-based model as the feature extractor. Based on the ego-motion estimation network of Monodepth2 \cite{mono}, Ozyoruk et al. \cite{endoslam} apply a spatial attention block into the feature extractor to perform ego-motion estimation for endoscope in the gastrointestinal tract. Similarly, Liu et al. \cite{liu} propose a dual-attention module in the feature encoder, in which spatial attention and channel attention are performed to extract contextual feature in both of the spatial dimension and the channel dimension. Yang et al. \cite{yang} combine CNN layers of ResNet with multi-head attention layers to perform feature extraction based on laparoscopic images. In the proposed joint feature extractor, a novel attention-based module is designed to extract and integrate both of local and global feature from multiple dimensions.


Moreover, in the most of the previous researches, the ego-motion estimation is performed on one certain dataset, such as the simulated dataset of the bronchoscopy \cite{offset},\cite{borrego}, the dataset of the laparoscopy \cite{liu},\cite{simcol},\cite{endoslam},\cite{yang}. The proposed framework is applied on three different datasets including an in-vivo dataset of the nasal endoscopy collected by human experts, a simulated dataset of colonoscopy \cite{simcol} and an ex-vivo dataset of intestine endoscopy collected by a robot arm \cite{endoslam}.

\section{Proposed Method}
\begin{figure*}
    \centering
    \includegraphics[width=0.9\textwidth]{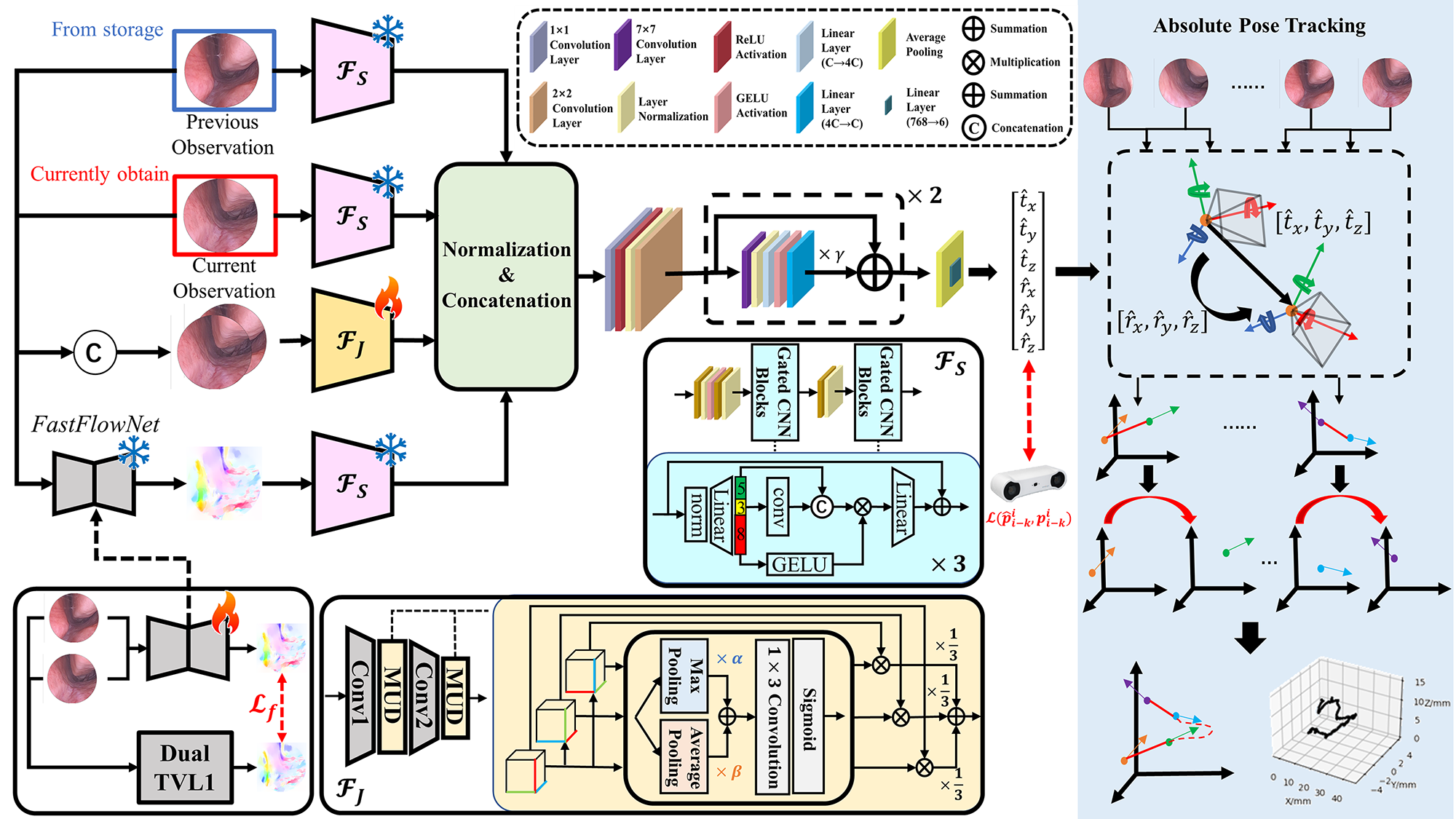}
    \caption{The pipeline of the proposed framework. $\mathcal{F}_{S}$ represents the feature extractor to extract scene features and the motion feature from two adjacent endoscopic images and the corresponding optical flow respectively. $\mathcal{F}_{J}$ represents the feature extractor to extract joint feature from the concatenation of two frames. In the pipeline of $\mathcal{F}_{J}$, 'Conv1' and 'Conv2' represent the first two layers of ResNet-34.}
    \label{pipeline}
\end{figure*}
\subsection{Overview}
\label{ov}
The real-time ego-motion tracking of the proposed method can be divided into two stage, relative pose estimation and absolute pose calculation, which is shown as Fig. \ref{pipeline}. At the first stage of relative pose estimation, given the current observation $I_{i}$ and the adjacent previous frame $I_{i-k}$, the relative pose transformation $p_{i-k}^{i}$ between two frames can be predicted by the proposed deep learning-based network. Based on a series of relative pose transformations $\{p_{(i-1)k}^{ik}\}_{i=1,2,...,N}$ and the known initial pose $P_0$, the absolute pose $P_{Nk}$ of the endoscope in the current frame can be calculated.

\subsection{Relative Pose Estimation}
\subsubsection{Optical Flow Prediction}
To achieve real-time tracking, the previous efficient network, FastFlowNet \cite{flow}, is utilized to predict the optical flow between two adjacent observations. However, the pretrained FastFlowNets are trained on the natural dataset, which may be influenced by the domain gap between the in-the-wild scene and the endoscopic scene. Therefore, the FastFlowNet in our framework is finetuned based on more accurate results calculated by a traditional but slow method. Given two adjacent frames $I_{i}$ and $I_{i-k}$, the optical flow $O_{i-k}^i$ as the ground truth is calculated by Dual TVL1 algorithm. Following \cite{flow}, the multi-scale robust loss between $O_{i-k}^i$ and the optical flow $\hat{O}_{i-k}^i$ predicted by the pretrained network is utilized for finetuning, formulated as Eq. \ref{lf}.
\begin{equation}
    \mathcal{L}_f = \sum_{l=2}^6 \theta_l \sum_{x}(|\hat{O}_{i-k}^i(x,l)-O_{i-k}^i(x,l)|+\epsilon)^q
\label{lf}
\end{equation}
where $l$ denotes the optical flow with different scales ($O_{i-k}^i(\cdot,l)\in \mathbb{R}^{H/2^{l-1}\times W/2^{l-1}\times 2}$), $x$ denotes each pixel in the optical flow, $\epsilon=0.01$ is a small constant of noise, $q<1$ is a penalty parameter which makes the model more robust for large magnitude outliers, and $\theta_l$ are set following \cite{flow}.
\subsubsection{Feature Extraction}
Firstly, the feature of the scenes $f_i$ and $f_{i-k}$ in two adjacent observations are extracted from single images respectively. Meanwhile, the feature of the endoscope motion $f_m$ is extracted from the optical flow of two adjacent observations by the same feature extractor. The network based on \cite{mo} is utilized as the feature extractor, which is pretrained on the Imagenet-1k \cite{imagenet}.

To extract the contextual and corresponding feature of two adjacent observations, the joint feature $f_j$ is extracted from the concatenation of two frames. Due to more information in the channel dimension of the concatenated image, a unique feature extractor is proposed for joint feature extraction. Specifically, based on the first two layers of ResNet-34, a attention-based module is inserted after each layer. Given the feature map $F_0$, the feature map $F$ is provided by integrating multi-dimensional information. Firstly, the $F_0$ is permuted into three feature maps with different dimensions, $F^0_0 \in \mathbb{R}^{H\times W\times C}$, $F^1_0 \in \mathbb{R}^{C\times H\times W}$ and $F^2_0 \in \mathbb{R}^{W\times C\times H}$. The attention weight of each feature map is calculated by Eq. \ref{fj}. To consider both of the critical information and global information, adaptive integration of max pooling $\mathcal{F}^{max}_{pooling}$ and average pooling $\mathcal{F}^{avg}_{pooling}$ is utilized to calculate attention weights, followed by a $1\times3$ convolution layer $\mathcal{F}^{1\times3}_{conv}$ and a sigmoid process. In the end, multiplications of the feature maps and corresponding attention weights are summed with equal weights as Eq. \ref{sum}.
\begin{equation}
    A^i = sigmoid(\mathcal{F}^{1\times3}_{conv}(\alpha \mathcal{F}^{max}_{pooling}(F^i_0) + \beta \mathcal{F}^{avg}_{pooling}(F^i_0)))
\label{fj}
\end{equation}
\begin{equation}
    F = \sum_{i=1}^3 \frac{1}{3}A^iF^i_0
\label{sum}
\end{equation}
while $\alpha$ and $\beta$ are two learnable weights, which are respectively set to a random value in the range of $[0,1)$. At the end of the feature extraction, all normalized feature maps are concatenated and passed through the proposed pose decoder as Eq. \ref{fusion} shows.
\begin{equation}
    f_{i-k}^i = f_i \oplus f_{i-k} \oplus f_m \oplus f_j
\label{fusion}
\end{equation}

\begin{table*}[]
\caption{The results of comparison with previous methods on three datasets}
\centering
\begin{tabular}{c|cc|ccccc}
\hline
\textbf{Dataset}          & \textbf{Methods} & \textbf{From}                             & \textbf{ATE/mm}    & \textbf{CE/e-3}    & \textbf{DE/deg}      & \textbf{RTE/mm}        & \textbf{ROT/deg}       \\ \hline
\multirow{8}{*}{NEPose}   & OffsetNet \cite{offset} & ICRA 2019            & 24.09±14.08        & 3.50±3.93          & 4.67±4.10            & 0.6427±0.3727          & 0.2160±0.2003          \\
                          & PoseResNet \cite{shao} & ICRA 2021             & 11.45±6.39         & 4.25±5.70          & 4.86±4.44            & 0.5506±0.3755          & 0.2389±0.2108          \\
                          & Attention PoseNet \cite{endoslam} & MedIA 2021 & 13.29±10.15        & 4.95±5.45          & 5.16±4.43            & 0.6386±0.4077          & 0.2413±0.2022          \\
                          & Fried et al. \cite{fried} & IROS 2023           & 11.77±8.77         & 3.71±4.09          & 4.90±3.97            & 0.5480±0.3471          & 0.2189±0.1956          \\
                          & PoseCorrNet \cite{simcol} & TMRB 2023          & 8.63±5.49          & 4.25±4.84          & 6.20±5.36            & 0.5110±0.3477          & 0.2233±0.1967          \\
                          & Dual-attention PoseNet \cite{liu} & CMPB 2023   & 14.37±10.23        & 3.64±4.31          & 4.92±4.25            & 0.6047±0.3857          & 0.2175±0.1922          \\
                          & Yang et al. \cite{yang} & TMI 2024             & 11.81±9.34         & 3.23±4.65          & 4.57±4.48            & 0.6050±0.4210          & 0.2166±0.1942          \\
                          &  \multicolumn{2}{c|}{Ours}                                       & \textbf{4.01±2.92} & \textbf{2.52±2.85} & \textbf{4.53±3.84}   & \textbf{0.4124±0.2757} & \textbf{0.2110±0.1797} \\ \hline
\textbf{Dataset}          & \textbf{Methods} & \textbf{From}                            & \textbf{ATE/cm}    & \textbf{CE/e-1}    & \textbf{DE/deg}      & \textbf{RTE/cm}        & \textbf{ROT/deg}       \\ \hline
\multirow{8}{*}{SimCol}   & OffsetNet \cite{offset} & ICRA 2019            & 8.02±4.56          & 2.45±2.88          & 29.49±14.48          & 0.4747±0.2864          & 1.7830±1.0802          \\
                          & PoseResNet \cite{shao} & ICRA 2021             & 6.26±4.45          & 1.97±2.34          & 26.81±16.61          & 0.4806±0.2876          & 1.3005±0.8255          \\
                          & Attention PoseNet \cite{endoslam} & MedIA 2021 & 7.21±4.99          & 2.18±2.38          & 29.85±17.03          & 0.5044±0.3024          & 1.4293±0.8902          \\
                          & Fried et al. \cite{fried} & IROS 2023           & 6.39±4.66          & 2.08±2.48          & 26.43±14.79          & 0.4735±0.2817          & 1.2877±0.8304          \\
                          & PoseCorrNet \cite{simcol} & TMRB 2023          & 6.08±4.87          & 1.97±2.62          & 27.73±15.53          & 0.4496±0.2489          & 1.2854±0.7640          \\
                          & Dual-attention PoseNet \cite{liu} & CMPB 2023   & 6.63±4.55          & 2.19±2.53          & 29.87±17.46          & 0.4848±0.2884          & 1.3541±0.8633          \\
                          & Yang et al. \cite{yang} & TMI 2024             & 10.27±7.36         & 3.73±3.59          & 32.08±19.17          & 0.7651±0.5573          & 2.2406±1.5303          \\
                          & \multicolumn{2}{c|}{Ours}                                      & \textbf{5.43±3.89} & \textbf{1.72±2.48} & \textbf{24.82±14.17} & \textbf{0.4472±0.2585} & \textbf{1.2258±0.7959} \\ \hline
\textbf{Dataset}          & \textbf{Methods}    & \textbf{From}                         & \textbf{ATE/cm}    & \textbf{CE/e-2}    & \textbf{DE/deg}      & \textbf{RTE/mm}        & \textbf{ROT/deg}       \\ \hline
\multirow{8}{*}{EndoSLAM} & OffsetNet \cite{offset} & ICRA 2019            & 10.04±5.68         & 3.64±4.08          & 19.51±13.79          & 5.07±5.47              & 1.1423±1.8140          \\
                          & PoseResNet \cite{shao} & ICRA 2021             & 12.20±7.18         & 4.58±5.25          & 22.04±15.30          & 5.23±5.56              & 1.2108±1.8297          \\
                          & Attention PoseNet \cite{endoslam} & MedIA 2021 & 11.14±6.40         & 3.31±3.99          & 16.91±11.22          & 4.93±5.56              & 1.1730±1.8069          \\
                          & Fried et al. \cite{fried} & IROS 2023           & 11.34±7.83         & 4.21±5.00          & 20.09±14.04          & 5.13±5.51              & 1.1814±1.8275          \\
                          & PoseCorrNet \cite{simcol} & TMRB 2023          & 10.54±5.29         & 4.47±5.02          & 22.37±14.82          & 5.08±5.52              & 1.2297±1.8252          \\
                          & Dual-attention PoseNet \cite{liu} & CMPB 2023   & 10.96±6.26         & 4.33±4.39          & 21.29±13.55          & 5.16±5.48              & 1.2115±1.8135          \\
                          & Yang et al. \cite{yang} & TMI 2024             & 10.82±5.74         & 4.36±7.45          & 19.83±18.11          & \textbf{4.84±5.44}     & 1.1853±1.8242          \\
                          & \multicolumn{2}{c|}{Ours}                                 & \textbf{9.86±4.28} & \textbf{2.30±2.89} & \textbf{14.59±10.34} & 4.93±5.43              & \textbf{1.1115±1.8264} \\ \hline
\end{tabular}
\label{comp}
\end{table*}

\subsubsection{Pose Decoder}
Based on the concatenated feature map $f_{i-k}^i$, the relative pose transformation vector $\hat{p}_{i-k}^i$ is predicted by a pose decoder, which can extract more complete representation from the feature map. At first, the feature map is squeezed by 
a $1\times 1$ convolution layer followed by a ReLU activation. After this, the feature map is downsampled by layer normalization and a $2\times 2$ convolution layer. For more complete extraction of feature representation, two blocks based on the depthwise separable convolution are used in our pose decoder. In each block, depthwise convolution is firstly performed by a $7\times 7$ convolution layer, in which the input feature map is divided into 3 groups for computation. Pointwise convolution is performed by two $1\times 1$ convolution layer to integrate the information in the feature maps from from different groups of the input. At the end of each block, the feature map is scaled by a learnable parameter $\gamma$ which is initialized as $10^{-6}$, followed by the residual connection with the input feature map. 

\subsection{Loss Function}
The predicted relative pose vector can be represented as $\hat{p}_{i-k}^i=[\hat{t}_{i-k}^i,\hat{q}_{i-k}^i]$, where $\hat{t}_{i-k}^i\in\mathbb{R}^3$ is the vector of translation and $\hat{q}_{i-k}^i\in\mathbb{R}^4$ is the quaternion of the endoscope. The ground truths of endoscope pose in two adjacent frames are $P_{i-k}$ and $P_i$. The relative pose transformation $P_{i-k}^i=P_{i-k}^{-1}P_i$ can be calculated, which can be transformed into $p_{i-k}^i=[t_{i-k}^i,q_{i-k}^i]$ based on the relation between the homogeneous matrix and the quaternion.

Refer to \cite{loss}, geometric loss function is utilized as our loss function. To calculate the loss function, the quaternion $q=[q^0,q^x,q^y,q^z]$ is represented as $q=[q^0,v]$. Based on this, the logarithm of the quaternion $\log q$ is defined as:
\begin{equation}
    \log q = 
    \left\{\begin{matrix}
    \frac{v}{\left\lVert v \right\rVert}\cos^{-1}q_0, & \left\lVert v \right\rVert \neq 0\\
    0, & \left\lVert v \right\rVert = 0
    \end{matrix}\right.
\end{equation}

Given the predicted relative pose vector $\hat{p}^{i}_{i-k}$ and the ground truth $p^{i}_{i-k}$ with $\lambda_1$ and $\lambda_2$ as two learnable parameters, the loss function $\mathcal{L}(\hat{p}_{i,i+k}, p_{i,i+k})$ is described as below:
\begin{equation}
\begin{split}
    \mathcal{L}(\hat{p}_{i-k}^i, p_{i-k}^i)=&\left\lVert t_{i-k}^i-\hat{t}_{i-k}^i \right\rVert_1e^{-\lambda_1}+\lambda_1+\\
    &\left\lVert \log q_{i-k}^i-\log \hat{q}_{i-k}^i \right\rVert_1e^{-\lambda_2}+\lambda_2
\end{split}
\end{equation}
where $\left\lVert \cdot \right\rVert_1$ represents the L1 norm, $\lambda_1$ and $\lambda_2$ is initially set to 0 and -3, respectively.

\subsection{Absolute Pose Calculation}
The predicted relative pose transformation $\hat{p}_{i-k}^i$ can be transformed into the corresponding homogeneous matrix $\hat{P}_{i-k}^i$. Given the know initial pose of the endoscope $\hat{P_0} = P_0$, the absolute pose of the endoscope can be calculated by the recursive formula Eq. \ref{abs} based on a series of relative pose transformation.
\begin{equation}
    \hat{P}_{i} = \hat{P}_{i-k}\hat{P}_{i-k}^i, i=k, 2k, ...
\label{abs}
\end{equation}

\section{Experiments and Results}

\begin{figure*}
    \centering
    \includegraphics[width=0.95\textwidth]{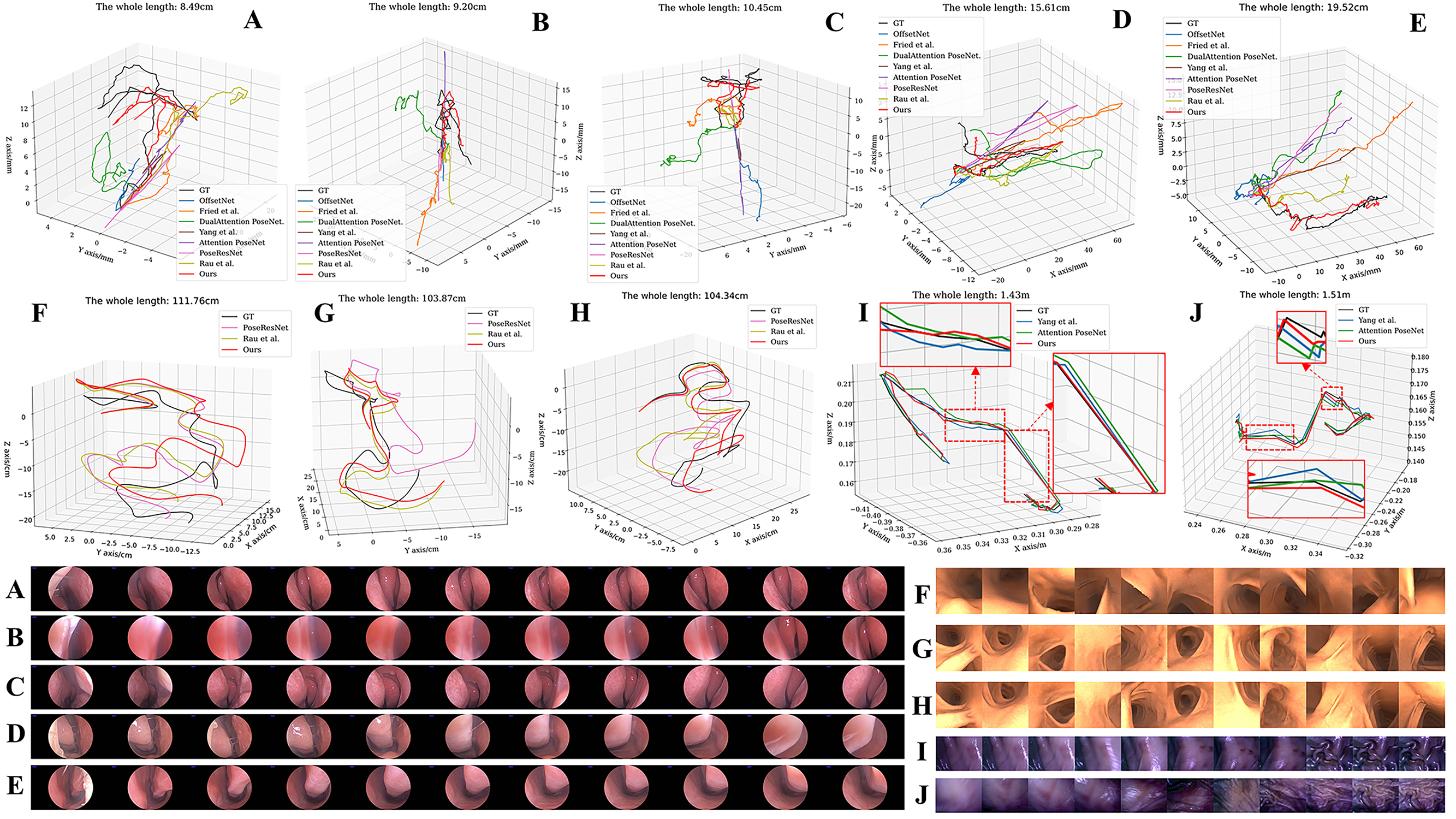}
    \caption{The trajectories tracking based on different methods. A-E are trajectories from dataset NEPose, F-H are trajectories from dataset SimCol (three best methods are compared for clear display), I and J are trajectories from dataset EndoSLAM. Specially, the trajectories in I and J are generated by $\hat{P_{i}}= P_{i-1}\hat{P}_{i-1}^i$ to evaluate the performance of relative pose estimation. The sampled frames of corresponding trajectories are shown at the bottom of the figure. The visualization of real-time position tracking can be found on the project page:\href{https://remote-bmxs.netlify.app}{remote-bmxs.netlify.app}.}
    \label{traj}
\end{figure*}
\begin{figure*}
    \centering
    \includegraphics[width=0.95\textwidth]{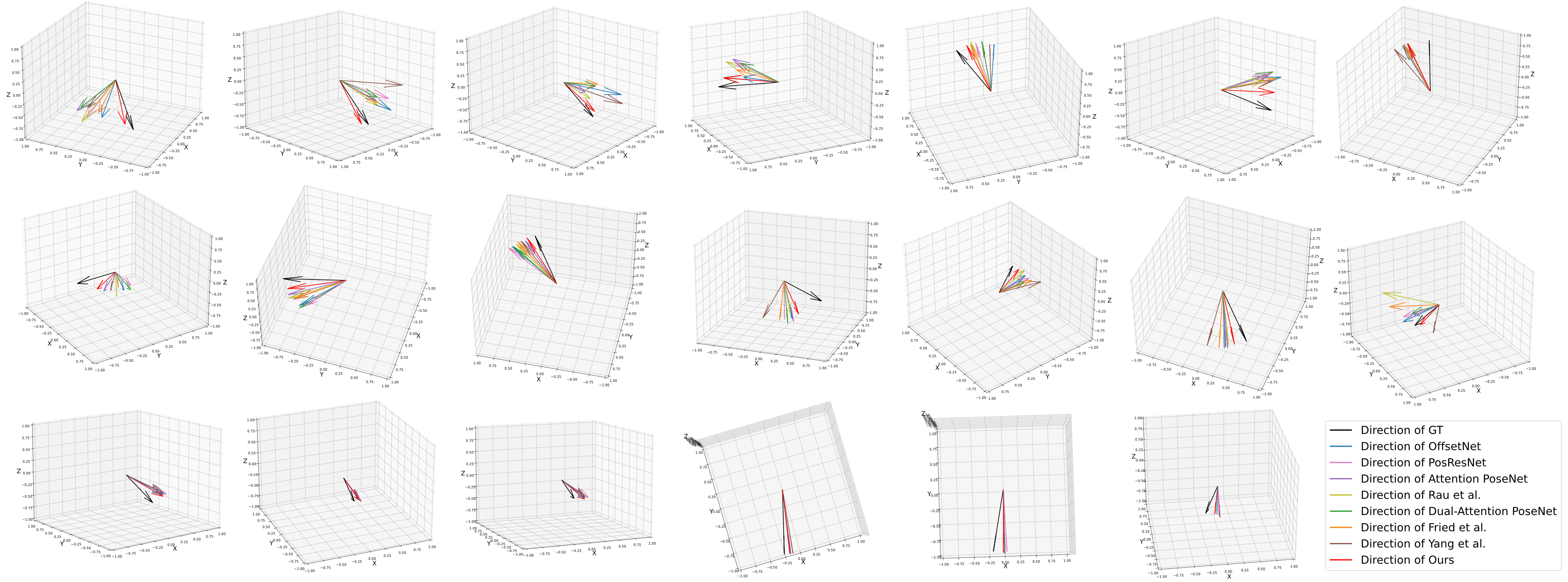}
    \caption{The visualization of directions predicted by different methods. \textbf{The first row}: results from SimCol dataset. \textbf{The second row}: results from EndoSLAM dataset. \textbf{The third row}: results from NEPose dataset. More examples can be found on the project page:\href{https://remote-bmxs.netlify.app}{remote-bmxs.netlify.app}.}
    \label{dirs}
\end{figure*}

\subsection{Datasets}
\subsubsection{NEPose}
NEPose is a dataset collected by our own. the complete process of nasal endoscopy is recorded by an experienced surgeon using a XION binocular 4K endoscope with an optical tracking system. The dataset consists of 50 endoscopic videos of 16 subjects consisting of over 30k frames. In the experiment, 4328 frames from 30 different subjects are randomly chosen as the training set and the validation set.
\subsubsection{SimCol}
SimCol \cite{simcol} is a virtual dataset generated based on a computer tomography scan of a real human colon in a Unity simulation environment. The dataset consists of 18k frames from 15 different subjects. In the experiment, 7200 frames from 24 different trajectories are randomly selected as the training set and the validation set.
\subsubsection{EndoSLAM}
EndoSLAM \cite{endoslam} is a large scale dataset for endoscopic perception. In our work, to perform ego-motion tracking, the ex-vivo part of EndoSLAM are utilized in the experiments. The chosen part of dataset consists of about 40k frames collected by a robot arm from ex-vivo colon, intestine and stomach. In the experiment, 3482 frames from 18 different trajectories are chosen from the dataset as the training set.

\subsection{Experiment Implementation}
\label{imp}
In the experiment, the proposed framework is implemented on Ubuntu 22.04 using Pytorch 1.13.1 with CUDA 11.7. Images are sampled every 4 frames from the video sequence, which means $k$ mentioned in \ref{ov} is set to 4. Our model is trained for 500 epochs with a batch size of 32 on one NVIDIA RTX 4090 GPU. Adam optimizer is used for optimization. The learning rate is set to 0.001 initially and decayed with a factor of 0.98 after each epoch.

\subsection{Metrics}
Absolute Translation Error (ATE) \cite{simcol} is used to evaluate the accuracy of absolute position tracking. Cosinus Error (CE) and Direction Error (DE) \cite{borrego} are used to evaluate the accuracy of absolute direction tracking. Moreover, Relative Translation Error (RTE) and ROTation error(ROT) \cite{simcol} are utilized to separately test the performance of relative translation prediction and relative rotation prediction.
\begin{equation}
     ATE = \left\lVert Trans(P_{ik})-Trans(\hat{P}_{ik}) \right\rVert
\end{equation}
\begin{equation}
     CE = \sum_{(\theta,\hat{\theta})\in\{(r_x,\hat{r}_x),(r_y,\hat{r}_y),(r_z,\hat{r}_z)\}}\frac{1}{3}(1-\cos{(\theta-\hat{\theta})})
\end{equation}
\begin{equation}
    DE = \cos^{-1}[(Rot\hat{P}_{ik})\cdot \left[\begin{matrix}
        1\\0\\0
    \end{matrix}\right])\cdot(Rot(P_{ik})\cdot \left[\begin{matrix}
        1\\0\\0
    \end{matrix}\right])]
\end{equation}
\begin{equation}
     RTE = \left\lVert Trans({P^{ik}_{(i-1)k}}^{-1}\hat{P}^{ik}_{(i-1)k}) \right\rVert
\end{equation}
\begin{equation}
     ROT = \frac{Tr(Rot({P^{ik}_{(i-1)k}}^{-1}\hat{P}^{ik}_{(i-1)k}))-1}{2}\cdot \frac{180}{\pi}
\end{equation}
where $Trans(P)$ represents the translation vector $T$ in the homogeneous matrix $P$, $Rot(P)$ represents the rotation matrix $R$ in the homogeneous matrix $P$, and $Tr(\cdot)$ represents the trace of the matrix.

\begin{table*}[]
\caption{Ablation on different modalities of inputs.}
\centering
\begin{tabular}{cccccccccccc}
\hline
\multicolumn{3}{c|}{\textbf{Modalities}}                           & \multicolumn{3}{c|}{\textbf{NEPose}}                                              & \multicolumn{3}{c|}{\textbf{SimCol}}                                                    & \multicolumn{3}{c}{\textbf{EndoSLAM}}                          \\
\textbf{S} & \textbf{C} & \multicolumn{1}{c|}{\textbf{O}} & \textbf{ATE/mm}    & \textbf{CE/e-3}    & \multicolumn{1}{c|}{\textbf{DE/deg}}    & \textbf{ATE/cm}    & \textbf{CE}            & \multicolumn{1}{c|}{\textbf{DE/deg}}      & \textbf{ATE/cm}     & \textbf{CE/e-2}    & \textbf{DE/deg}     \\ \hline
\CheckmarkBold             & \CheckmarkBold             & \multicolumn{1}{c|}{\CheckmarkBold}             & \textbf{4.95±3.82} & \textbf{4.70±6.98} & \multicolumn{1}{c|}{\textbf{4.67±3.98}} & \textbf{5.83±3.77} & \textbf{1.69±2.00} & \multicolumn{1}{c|}{\textbf{23.01±14.87}} & \textbf{10.41±4.19} & \textbf{1.81±2.41} & \textbf{12.56±8.01} \\
\XSolidBrush             & \CheckmarkBold             & \multicolumn{1}{c|}{\CheckmarkBold}             & 11.82±8.00         & 5.39±7.51          & \multicolumn{1}{c|}{5.46±4.87}          & 6.84±4.43          & 1.89±2.33         & \multicolumn{1}{c|}{23.29±11.92}          & 11.17±5.46          & 2.35±2.71          & 15.02±9.85          \\
\CheckmarkBold             & \XSolidBrush             & \multicolumn{1}{c|}{\CheckmarkBold}             & 6.13±3.80          & 4.72±7.53          & \multicolumn{1}{c|}{4.75±3.89}          & 6.40±4.53          & 1.71±1.94          & \multicolumn{1}{c|}{23.91±14.82}          & 12.96±6.79          & 1.82±2.40          & 12.61±8.31          \\
\CheckmarkBold             & \CheckmarkBold             & \multicolumn{1}{c|}{\XSolidBrush}             & 6.80±4.40          & 5.62±8.48          & \multicolumn{1}{c|}{5.10±4.59}          & 5.86±3.81          & 1.74±1.97          & \multicolumn{1}{c|}{25.70±15.46}          & 11.91±5.87          & 1.88±2.38          & 13.31±7.74          \\
\XSolidBrush             & \XSolidBrush             & \multicolumn{1}{c|}{\CheckmarkBold}             & 16.00±11.78        & 5.74±7.90          & \multicolumn{1}{c|}{5.59±4.93}          & 10.79±5.53         & 2.96±3.28         & \multicolumn{1}{c|}{41.53±16.59}          & 10.80±4.36          & 1.96±2.45          & 13.09±9.37          \\
\CheckmarkBold             & \XSolidBrush             & \multicolumn{1}{c|}{\XSolidBrush}             & 5.77±3.75          & 4.80±6.34          & \multicolumn{1}{c|}{5.30±4.35}          & 6.07±3.91          & 1.87±2.02          & \multicolumn{1}{c|}{26.89±15.25}          & 10.45±4.08          & 2.10±2.46          & 14.01±9.05          \\
\XSolidBrush             & \CheckmarkBold             & \multicolumn{1}{c|}{\XSolidBrush}             & 9.55±6.21          & 5.53±7.40          & \multicolumn{1}{c|}{5.15±4.78}          & 6.25±4.07          & 1.91±1.91          & \multicolumn{1}{c|}{28.28±12.97}          & 12.99±6.51          & 3.56±3.50          & 19.05±11.05         \\ \hline
\multicolumn{12}{l}{\textbf{S} denotes two separate adjacent observations, \textbf{C} denotes the concatenation of two frames, and \textbf{O} denotes the optical flow.}        
\end{tabular}
\label{ab1}
\end{table*}

\begin{table*}[]
\caption{Ablation on different feature extractors and pose decoders.}
\centering
\begin{tabular}{cccccccccccc}
\hline
\multirow{2}{*}{\textbf{$\mathcal{F}_S$}} & \multirow{2}{*}{\textbf{$\mathcal{F}_J$}} & \multicolumn{1}{c|}{\multirow{2}{*}{\textbf{$\mathcal{F}_P$}}} & \multicolumn{3}{c|}{\textbf{NEPose}}                                        & \multicolumn{3}{c|}{\textbf{SimCol}}                                            & \multicolumn{3}{c}{\textbf{EndoSLAM}}                     \\
                                          &                                           & \multicolumn{1}{c|}{}                                          & \textbf{ATE/mm}  & \textbf{CE/e-3}  & \multicolumn{1}{c|}{\textbf{DE/deg}}  & \textbf{ATE/cm}  & \textbf{CE/e-1}    & \multicolumn{1}{c|}{\textbf{DE/deg}}    & \textbf{ATE/cm}    & \textbf{CE/e-2}  & \textbf{DE/deg}   \\ \hline
Ours                                      & Ours                                      & \multicolumn{1}{c|}{Ours}                                      & \textbf{4.0±2.9} & \textbf{2.6±2.9} & \multicolumn{1}{c|}{\textbf{4.6±3.9}} & \textbf{4.6±2.9} & \textbf{1.60±2.61} & \multicolumn{1}{c|}{\textbf{21.8±10.5}} & \textbf{8.59±3.96} & \textbf{2.2±2.8} & \textbf{14.3±9.8} \\
Res                                       & Ours                                      & \multicolumn{1}{c|}{Ours}                                      & 8.9±5.2          & 3.4±3.7          & \multicolumn{1}{c|}{5.1±4.3}          & 5.4±3.7          & 1.73±2.12          & \multicolumn{1}{c|}{27.8±15.6}          & 9.06±6.00          & 3.5±3.4          & 19.7±12.6         \\
Ours                                      & w/o MUD                                   & \multicolumn{1}{c|}{Ours}                                      & 4.8±3.4          & 2.8±3.0          & \multicolumn{1}{c|}{4.7±3.8}          & 4.8±3.3          & 1.63±2.60          & \multicolumn{1}{c|}{22.3±9.5}           & 8.62±5.03          & 2.4±2.8          & 14.6±10.2         \\
Ours                                      & Ours                                      & \multicolumn{1}{c|}{Pre}                                       & 8.5±5.9          & 3.3±3.9          & \multicolumn{1}{c|}{5.5±4.6}          & 5.3±3.7          & 1.68±2.53          & \multicolumn{1}{c|}{26.6±15.4}          & 8.95±4.68          & 2.9±3.1          & 17.3±11.3         \\
Ours                                      & Ours                                      & \multicolumn{1}{c|}{FC}                                        & 12.2±7.1         & 3.1±3.2          & \multicolumn{1}{c|}{5.1±4.1}          & 5.9±3.2          & 1.68±2.60          & \multicolumn{1}{c|}{23.3±11.4}          & 8.68±4.03          & 2.4±3.0          & 14.9±10.4         \\ \hline
\multicolumn{12}{l}{\begin{tabular}[c]{@{}l@{}}$\mathcal{F}_S$,$\mathcal{F}_J$ and 'MUD' are modules described in Fig. \ref{pipeline}, $\mathcal{F}_P$ denotes the pose decoder in the framework, 'FC' denotes the fully-connected layer, \\'Res' denotes the pretrained ResNet-18 used in \cite{simcol},\cite{shao}, 'Pre' denotes the pose decoder used in previous works including \cite{simcol},\cite{mono},\cite{endoslam}.\end{tabular}}              
\end{tabular}
\label{ab2}
\end{table*}

\subsection{Comparison with Previous Methods}
The proposed method is compared with state-of-the-art methods including OffsetNet\cite{offset}, PoseResNet\cite{shao}, Attention PoseNet\cite{endoslam}, the method from \cite{fried}, the method from \cite{simcol}, Dual-attention PoseNet \cite{liu} and the most recent method from \cite{yang}. For comparison, 1308 frames from 12 different subjects in NEPose dataset, 1800 frames from 6 different trajectories in SimCol dataset and 1482 frames from 6 different trajectories in EndoSLAM dataset are randomly selected in the experiments. The same implementation described in \ref{imp} is used in the experiments of all methods. The experimental results on three datasets are respectively shown as Table \ref{comp}. The results demonstrate that the proposed method can perform the most accurate relative pose estimation and ego-motion tracking for various endoscopic scenes, compared with previous works. Fig. \ref{traj} and Fig. \ref{dirs} visualize the results of ego-motion tracking, which displays the best qualitative results are from the proposed method.

\subsection{Ablation Studies} 
In the ablation studies, we set several experiments to prove the effect of each visual modal and the effect of novel modules, which includes feature extractors and the pose decoder. In the experiments, 1297 frames from NEPose, 1800 frames from SimCol and 1050 frames from EndoSLAM are randomly selected again for the comparison. The results in Table \ref{ab1} demonstrate that each visual modal plays an important role in the ego-motion tracking by our framework. To evaluate the positive effects of the proposed modules, the feature extractor is compared with pretrained ResNet-18, which is usually used in the previous work. Meanwhile, the pose decoder is replaced with the advanced pose decoder and a fully-connect layer, which are usually applied in the previous framework. In the experiments, 1279 frames from NEPose, 1800 frames from SimCol and 1145 frames from EndoSLAM are randomly selected again for the comparison. The results in Table \ref{ab2} prove the effects of the novel feature extractors and the proposed pose decoder.

\subsection{Inference Speed Test}
To evaluate the inference speed of the proposed framework, 18 samples are randomly selected from all three datasets for experiments. As Table \ref{spd} shows, the average inference speed of our framework can be over 30 frames per second (fps), which meets real-time requirement for application.

\begin{table}[]
\caption{Experimental results of inference speed on three datasets}
\centering
\begin{tabular}{c|cc|cc|cc}
\hline
\multirow{2}{*}{\textbf{\#}} & \multicolumn{2}{c|}{\textbf{NEPose}}                                            & \multicolumn{2}{c|}{\textbf{SimCol}}                                            & \multicolumn{2}{c}{\textbf{EndoSLAM}}                                           \\
                             & \textbf{\begin{tabular}[c]{@{}c@{}}Length\\ (mm)\end{tabular}} & \textbf{FPS}   & \textbf{\begin{tabular}[c]{@{}c@{}}Length\\ (cm)\end{tabular}} & \textbf{FPS}   & \textbf{\begin{tabular}[c]{@{}c@{}}Length\\ (cm)\end{tabular}} & \textbf{FPS}   \\ \hline
1                            & 119.3                                                          & \textbf{25.63} & 104.3                                                          & \textbf{34.18} & 145.4                                                          & \textbf{100.7} \\
2                            & 82.7                                                           & \textbf{22.89} & 102.2                                                          & \textbf{33.85} & 133.4                                                          & \textbf{98.3}  \\
3                            & 73.9                                                           & \textbf{40.83} & 103.8                                                          & \textbf{35.64} & 79.5                                                           & \textbf{86.4}  \\
4                            & 154.9                                                          & \textbf{33.86} & 111.5                                                          & \textbf{31.58} & 90.3                                                           & \textbf{90.5}  \\
5                            & 151.8                                                          & \textbf{30.21} & 111.6                                                          & \textbf{32.38} & 150.1                                                          & \textbf{86.75} \\
6                            & 174.3                                                          & \textbf{32.17} & 111.7                                                          & \textbf{31.35} & 145.4                                                          & \textbf{99.5}  \\ \hline
\textbf{Avg.}                & \multicolumn{2}{c|}{\textbf{31 fps}}                                            & \multicolumn{2}{c|}{\textbf{33 fps}}                                            & \multicolumn{2}{c}{\textbf{94 fps}}                                             \\ \hline
\end{tabular}
\label{spd}
\end{table}

\section{Conclusion}
In this paper, a deep learning-based framework, REMOTE, is proposed to perform real-time ego-motion tracking for endoscope in various scenes, in which different aspects of features from multi-modal visual inputs are learned. The experimental results show that the proposed method outperforms the advanced methods on three different endoscopic datasets, providing the most accurate real-time ego-motion estimation. This work could contribute to the automation and navigation of robot-assisted endoscopy.

\bibliographystyle{IEEEtran}
\bibliography{references}

\end{document}